\documentclass[runningheads]{llncs}
\usepackage{graphicx}
\usepackage{amsmath,amssymb} 
\usepackage{color}
\usepackage[width=122mm,left=12mm,paperwidth=146mm,height=193mm,top=12mm,paperheight=217mm]{geometry}
\usepackage{times}
\usepackage{epsfig}
\usepackage{graphicx}
\usepackage{amsmath}
\usepackage{amssymb}
\usepackage{amssymb}
\usepackage{multirow}
\usepackage{array}
\usepackage{booktabs}
\usepackage{tabularx}
\usepackage{algorithm}
\usepackage{algorithmic}
\usepackage{stmaryrd}
\usepackage{color}
\usepackage{makecell}
\usepackage{subfigure}

\begin{document}
\pagestyle{headings}
\mainmatter
\def\ECCV18SubNumber{559}  

\title{A Modulation Module for Multi-task Learning with Applications in Image Retrieval} 



\authorrunning{X. Zhao, H. Li, X. Shen, X. Liang, Y. Wu}

\author{Xiangyun Zhao$^{1}$\thanks{Part of the work is done when Xiangyun Zhao was an intern at Adobe Research advised by Haoxiang Li and Xiaohui Shen.} \quad Haoxiang Li$^{2}$ \quad Xiaohui Shen$^{3}$ \quad Xiaodan Liang$^{4}$ \quad Ying Wu$^{1}$}

\institute{$^{1}$  Northwestern University   $^{2}$ AIBee \\   $^{3}$ Bytedance AI Lab   $^{4}$Carnegie Mellon University  
}

\institute{ EECS Department, Northwestern University \and AIBee  \and Bytedance AI Lab \and Carnegie Mellon University }

\maketitle

\begin{abstract}

Multi-task learning has been widely adopted in many computer vision tasks to improve overall computation efficiency or boost the performance of individual tasks, under the assumption that those tasks are correlated and complementary to each other. However, the relationships between the tasks are complicated in practice, especially when the number of involved tasks scales up. When two tasks are of weak relevance, they may compete or even distract each other during joint training of shared parameters, and as a consequence undermine the learning of all the tasks. 
This will raise \textit{destructive interference} which decreases learning efficiency of shared parameters and lead to low quality loss local optimum w.r.t. shared parameters. To address the this problem, we propose a general modulation module, which can be inserted into any convolutional neural network architecture, to encourage the coupling and feature sharing of relevant tasks while disentangling the learning of irrelevant tasks with minor parameters addition. Equipped with this module, gradient directions from different tasks can be enforced to be consistent for those shared parameters, which benefits multi-task joint training. The module is end-to-end learnable without ad-hoc design for specific tasks, and can naturally handle many tasks at the same time. We apply our approach on two retrieval tasks, face retrieval on the CelebA dataset~\cite{liu2015faceattributes} and product retrieval on the UT-Zappos50K dataset~\cite{finegrained,semjitter}, and demonstrate its advantage over other multi-task learning methods in both accuracy and storage efficiency.  Code will be releassed here https://github.com/Zhaoxiangyun/Multi-Task-Modulation-Module.
\end{abstract}

\section{Introduction}

\begin{figure}[!t]
\centering
\includegraphics[scale = 0.2]{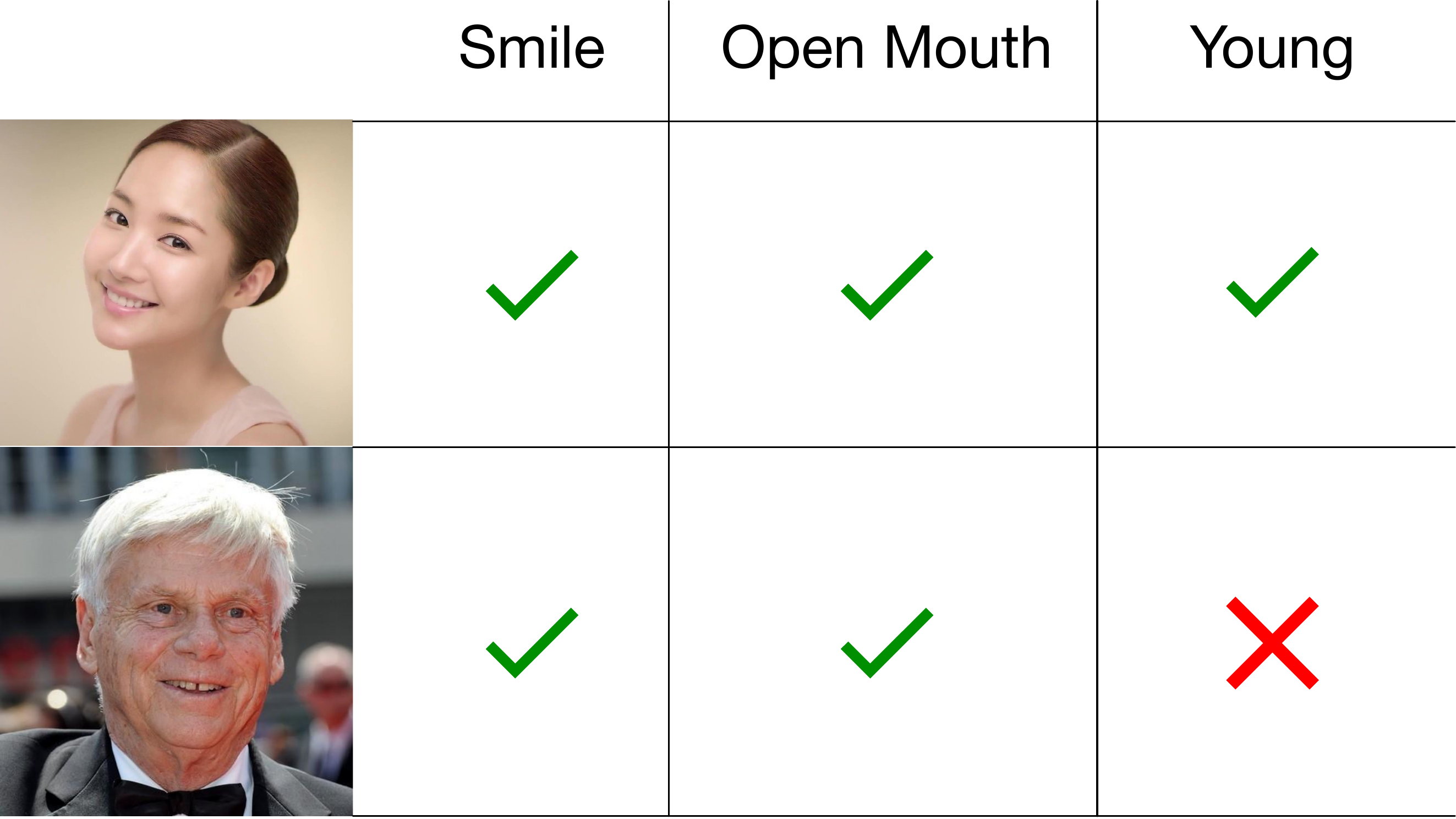}
\label{fig.motivation}
\caption{Conflicting training signals in multi-task learning: when jointly learning discriminative features for multiple face attributes, some samples may introduce conflicting training signals in updating shared model parameters, such as ``Smile" vs. ``Young".}
\end{figure}

Multi-task learning aims to improve learning efficiency and boost the performance of individual tasks by jointly learning multiple tasks at the same time. With the recent prevalence of deep learning-based approaches in various computer vision tasks, multi-task learning is often implemented as parameter sharing in certain intermediate layers in a unified convolutional neural network architecture~\cite{yin2017multi,ranjan2017all}. However, such feature sharing only works when the tasks are correlated and complementary to each other. When two tasks are irrelevant, they may provide competing or even contradicting gradient directions during feature learning. For example, learning to predict face attributes of ``Open Mouth'' and ``Young'' can lead to discrepant gradient directions for the examples in Figure~\ref{fig.motivation}. Because the network is supervised to produce nearby embeddings in one task but faraway embeddings in the other task, the shared parameters get conflicting training signals. It is analogous to the \textit{destructive interference} problem in Physics where two waves of equal frequency and opposite phases cancel each other. It would make the joint training much more difficult and negatively impact the performance of all the tasks. 

Although this problem is rarely identified in the literature, many of the existing methods are in fact designed to mitigate \textit{destructive interference} in multi-task learning. For example, in the popular multi-branch neural network architecture and its variants, the task-specific branches are designed carefully with the prior knowledge regarding the relationships of certain tasks~\cite{ranjan2016hyperface,jou2016deep,rothe2015dex}. By doing this, people expect less conflicting training signals to the shared parameters. Nevertheless, it is difficult to generalize those specific designs to other tasks where the relationships may vary, or to scale up to more tasks such as classifying more than 20 facial attributes at the same time, where the task relationships become more complicated and less well studied.


To overcome these limitations, we propose a novel modulation module, which can be inserted into arbitrary network architecture and learned through end-to-end training. It can encourage correlated tasks to share more features, and at the same time disentangle the feature learning of irrelevant tasks. In back-propagation of the training signals, it modulates the gradient directions from different tasks to be more consistent for those shared parameters; in the feed-forward pass, it modulates the features towards task-specific feature spaces. Since it does not require prior knowledge of the relationships of the tasks, it can be applied to various multi-task learning problems, and handle many tasks at the same time. One related work is \cite{shazeer2017outrageously} which try to increase model capacity without a proportional increase in computation.


To validate the effectiveness of the proposed approach, we apply the modulation module in a neural network to learn the feature embedding of multiple attributes, and evaluate the learned feature representations on diverse retrieval tasks. In particular, we first propose a joint training framework with several embedded modulation modules for the learning of multiple face attributes, and evaluate the attribute-specific face retrieval results on the CelebA dataset. In addition, we provide thorough analysis on the task relationships and the capability of the proposed module in promoting correlated tasks while decoupling unrelated tasks. Experimental results show that the advantage of our approach is more significant with more tasks involved, showing its generalization capability to larger-scale multi-task learning problems. Compared with existing multi-task learning methods, the proposed module learns improved task-specific features and supports a compact model for scalability. We further apply the proposed approach in product retrieval on the UT-Zappos50K dataset, and demonstrate its superiority over other state-of-the-art methods.

Overall, the contributions of this work are four-fold:

\begin{itemize}
\item We address the \textit{destructive interference} problem of unrelated tasks in multi-task learning, which is rarely discussed in previous work.
\item We propose a novel modulation module that is general and end-to-end learnable, to adaptively couple correlated tasks while decoupling unrelated ones during feature learning. 
\item With minor task-specific overhead, our method supports scalable multi-task learning without manually grouping of tasks.
\item We apply the module to the feature learning of multiple attributes, and demonstrate its effectiveness on retrieval tasks, especially on large-scale problems (e.g., as many as 20 attributes are jointly learned).
\end{itemize}
\section{Related Work}

\subsection{Multi-task learning}

It has been observed in many prior works that jointly learning of multiple correlated tasks can help improve the performance of each of them, for example, learning face detection with face alignment~\cite{ranjan2017all,zhang2016joint}, learning object detection with segmentation~\cite{He_2017_ICCV,girshick14CVPR}, and learning semantic segmentation with depth estimation~\cite{mousavian2016joint,wang2015towards}. While these works mainly study what related tasks can be jointly learned in order to mutually benefit each other, we instead investigate a proper joint training scheme given any tasks without assumption on their relationships.

A number of research efforts have been devoted to exploiting the correlations among related tasks for joint training. For example, Jou et al.~\cite{jou2016deep} propose the Deep Cross Residual Learning to introduce the cross-residuals connections as a form of network regularization for better network generalization. Misra et al.~\cite{misra2016cross} propose the Cross-stitch Networks to combine the activations from multiple task-specific networks for better joint training. Kokkinos et al.~\cite{KokkinosCVPR2017} propose UberNet to jointly learn low-, mid-, and high-level vision tasks by branching out task-specific paths from different stages in a deep CNN.

Most multi-task learning frameworks, if not all, involve parameters shared across tasks and task-specific parameters. In joint learning beyond similar tasks, it is desirable to automatically discover what and how to share between tasks. Recent works along this line include Lu et al.~\cite{Lu_2017_CVPR}, who propose to automatically discover a neural network design to group similar tasks together; Yang et al.~\cite{YangH16}, who model this problem as tensor factorization to learn how to share knowledge across tasks; and Veit et al.~\cite{Veit2017}, who propose to share all neural network layers but masking the final image features differently conditioned on the attributes/tasks.

Compared to these existing works, in this paper, we explicitly identify the problem of \textit{destructive interference} and propose a metric to quantify it. Our observation further confirms its correlation to the quality of learned features. Moreover, our proposed module is end-to-end learnable and flexible to be inserted anywhere into an existing network architecture. Hence, our method can further enhance the structure learned with the algorithm from Lu et al.~\cite{Lu_2017_CVPR} to improve its suboptimal within-group branches. When compared with the tensor factorization by Yang et al.~\cite{YangH16}, our module is lightweight, easy to train, and with a small and accountable overhead to include additional tasks. Condition similar networks~\cite{Veit2017} shares this desirable scalability feature with our method in storage efficiency. However, as they do not account for the \textit{destructive interference} problem in layers other than the final feature layer, we empirically observe that their method does not scale-up well in accuracy for many tasks (See Section~\ref{sec:faceresults}).




\subsection{Image Retrieval}

In this work, we evaluate our method with applications on image retrieval. Image retrieval has been widely studied in computer vision~\cite{philbin2007object,shen2012object,wan2014deep,wang2010semi,jegou2011product,perronnin2010large}. We do not study the efficiency problem in image retrieval as in many prior works~\cite{wang2010semi,Lin_2015_CVPR_Workshops,jegou2011product,perronnin2010large}. Instead, we focus on learning discriminative task-specific image features for accurate retrieval.


Essentially, our method is related to how discriminative image features can be extracted. In the era of deep learning, feature extraction is a very important and fundamental research direction. From the early pioneering AlexNet~\cite{krizhevsky2012imagenet} to recent seminal ResNet~\cite{he2016deep} and DenseNet~\cite{Huang_2017_CVPR}, the effectiveness and efficiency of neural networks have been largely improved. This line of research focuses on designing better neural network architectures, which is independent of our method. By design, our algorithm can potentially benefit from better backbone architectures.

Another important related research area is metric learning~\cite{xing2003distance,schapire2000boostexter,weinberger2009distance,schultz2004learning}, which mostly focuses on designing an optimization objective to find a metric to maximize the inter-class distance while minimizing the intra-class distance. They are often equivalent to learning a discriminative subspace or feature embedding. Some of them have been introduced into deep learning as the loss function for better feature learning~\cite{schroff2015facenet,hadsell2006dimensionality}. Our method is by design agnostic to the loss function, and we can potentially benefit from more sophisticated loss functions to learn more discriminative image feature for all tasks. In our experiment, we use triplet loss~\cite{schroff2015facenet} due to its simplicity.

\section{Our Method}

In this section, we first identify the  \textit{destructive interference} problem in sharing features for multi-task learning and then present the technical details of our modulation module to resolve this problem.

\begin{figure*}[!t]
\centering
\includegraphics[width=\linewidth]{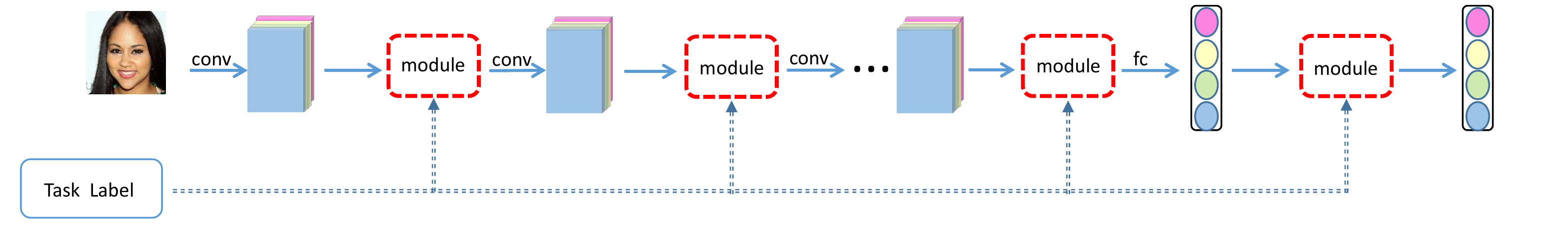}
\caption{A neural network fully modulated by our proposed modules: in testing, the network takes inputs as the image and task label to extract discriminative image features for the specified task.}
\label{fig.network1}
\end{figure*}



\renewcommand{\arraystretch}{1.2}
\begin{table}
    \begin{center}
     \scalebox{0.79}{   \begin{tabular}{|l|c|c|c|c|c|}
            \hline
             & smile Acc. & open mouth Acc.& young Acc.& smile / young UCR & smile /open-mouth UCR\\
            \hline\hline
            smile + young + open mouth(a) & 84.71\% &74.73 \% & 71.6\% &  -& - \\
            \hline
            smile + young(b) &  83.85\%& - &74.71\% &22.1\% & -\\
            \hline
            smile + open mouth(c) &  91.72\%&  92.65\%& - &- & 43.71\% \\
            \hline
            Three Independent Networks(d)  &  93.32\%& 94.40\% & 84.90\% & - & -\\
            \hline
            With Proposed Modulation(e) & 94.03\% & 95.31\%  & 86.20\% & \textbf{50.63\%} & \textbf{52.77\%}\\
            \hline
            With Proposed Modulation + Reg(f) & \textbf{94.94\%} & \textbf{95.58\% } & \textbf{87.75\% } & - & - \\
            \hline
        \end{tabular}}
    \end{center}
    \caption{Accuracy and UCR Comparison on three face attribute-based retrieval tasks (See Section~\ref{sec:expsetup} for details): the  comparison empirically support our analysis of the {\em destructive interference} problem and the assumption that reasonable task-specific modulation parameters can be learned from data}
    \label{tab:three}
\end{table}

\subsection{Destructive interference}

Despite that a multi-task neural network can have many variants which involve the learning of different task combinations, the fundamental technique is to share intermediate network parameters for different tasks,  and jointly train with all supervision signals from different tasks by gradient descent methods. One issue raised from this common scheme is that two irrelevant or weakly relevant tasks may drag gradients propagated from different tasks in conflicting or even opposite directions. Thus, learning the shared parameters can suffer from the well-known {\em destructive interference} problem.




Formally, we denote $\theta$ as the parameters of a neural network $F$ over different tasks, $I$ as its input, and $f = F(I|\theta)$ as its output. The update of $\theta$ follows its gradient: 
\begin{equation} 
\nabla \theta = \frac{\partial L}{\partial f}\frac{\partial f}{\partial \theta},
\end{equation} where $L$ is the loss function.

In multi-task learning, $\theta$ will be updated by gradients from different tasks. Essentially, $\frac{\partial L}{\partial f}$ directs the learning of $\theta$. In common cases, a discriminative loss generally encourages $f_i$ and $f_j$ to be similar for images $I_i$ and $I_j$ from the same class. However, the relationship of $I_i$ and $I_j$ can change in multi-task learning, even flip in different tasks. When training all these tasks, the update directions of $\theta$ may be conflicting, which is the namely {\em destructive interference} problem.

More specifically, given a mini-batch of training samples from task $t$ and $t^\prime$, $\nabla \theta = \nabla \theta_{t} + \nabla \theta_{t^\prime}$, where $\nabla \theta_{t/t^\prime}$ denotes gradients from samples of task $t/t^\prime$. Gradients from two tasks are negatively impacting the learning of each other, when 
\begin{equation}
A_{t,t^\prime} = sign(\langle \nabla \theta_{t}, \nabla \theta_{t^\prime}\rangle) = -1.
\label{eq:AngleOfTwoTasks}
\end{equation}

The {\em destructive interference} hinders the learning of the shared parameters and essentially leads to low quality loss local optimum w.r.t. shared parameters.

\subsubsection{Empirical Evidence~\label{sec:Destructive-exp}}

We validate our assumption through a toy experiment on jointly learning of multiple attribute-based face retrieval tasks. More details on the experimental settings can be found in Section~\ref{sec:expsetup}. 

Intuitively, the attribute {\em smile} is related to attribute {\em open mouth} but irrelevant to attribute {\em young}~\footnote{Here the attribute refers to its estimation from a given face image.}. As shown in Table~\ref{tab:three}, when we share all the parameters of the neural network across different tasks, the results degrade when jointly training the tasks compared with training three independent task-specific networks. The degradation when jointly training  {\em smile} and {\em young} is much more significant than the one when jointly training  {\em smile} and {\em open mouth}. That is because there are always some conflicting gradients from some training samples even if two tasks are correlated, and apparently when the two tasks are with weak relevance, the conflicts become more frequent, making the joint training ineffective. 

To further understand how the learning leads to the above results, we follow Equation~\ref{eq:AngleOfTwoTasks} to quantitatively estimate the compatibility of task pairs by looking at the ratio of mini-batches with $A_{t,t^\prime} > 0$ in one training epoch. So we define this ratio as Update Compliance Ratio(UCR) which measures the consistence of two tasks. The larger the UCR is, the more consistent the two tasks are in joint training. As shown in Table~\ref{tab:three}, in joint learning of {\em smile} and {\em open mouth} we observe higher compatibility compared with joint learning of {\em smile} and {\em young}, which explains the accuracy discrepancy from (b) to (c) in Table~\ref{tab:three}. Comparing (e) with (b) and (c), the accuracy improvement is accompanied with UCR improvement which explains how the proposed module improves the overall performance. With our proposed method introduced as following, we observe increased UCR for both task pairs.


\subsection{A Modulation Module~\label{sec:GM}}

Most multi-task learning frameworks involve task-specific parameters and shared parameters. Here we introduce a modulation module as a generic framework to add task-specific parameters and link it to alleviation of {\em destructive interference}.


More specifically, we propose to modulate the feature maps with task-specific projection matrix ${\mathbf W}_t$ for task $t$. As illustrated in Figure~\ref{fig.network1}, this module maintains the feature map size to keep it compatible with layers downwards in the network architecture. Following we will discuss how this design affects the back-propagation and feed-forward pass.


\subsubsection{Back-propagation}
In back-propagation, {\em destructive interference} happens when gradients from two tasks $t$ and $t^\prime$ over the shared parameters $\theta$ have components in conflicting directions, i.e., $\langle \nabla \theta_{t}, \nabla \theta_{t^\prime}\rangle < 0$. It can be simply derived that the proposed modulation over feature maps is equivalent to modulating shared parameters with task-specific masks ${\mathbf M}_{t/t^\prime}$. With the proposed modulation, the update to $\theta$ is now ${\mathbf M}_t\nabla \theta_t + {\mathbf M}_{t^\prime} \nabla \theta_{t^\prime}$. Since the task-specific masks/projection matrices are learnable, we observe that the training process will naturally mitigate the {\em destructive interference} by reducing the average across-task gradient angles $\langle {\mathbf M}_t \nabla \theta_{t}, {\mathbf M}_{t^\prime}\nabla \theta_{t^\prime}\rangle$, which is observed to result in better local optimum of shared parameters.

\subsubsection{Feed-Forward Pass~\label{subsec:FP}}

Given feature map $x$ with size $M \times N \times C$ and the modulation projection matrix ${\mathbf W}$, we have
\begin{equation} 
x' = {\mathbf W}_t \times x,
\end{equation} which is the input to the next layer.

A full projection matrix would require ${\mathbf W}_t$ of size $MNC \times MNC$, which is infeasible in practice and the modulation would degenerate to completely separated branches with a full project matrix. Therefore, we firstly simplify the $W_t$ to have shared elements within each channel. Formally, ${\mathbf W} = \{w_{i,j}\}, \{i,j\} \in \{1,\ldots,C\}$
\begin{equation}
x^\prime_{mni} = \sum_{j=1}^{C} x_{mnj}*w_{i,j},
\end{equation} where $x^\prime_{mni}$, $x_{mni}$ and $w_{ij}$  denote elements from input, output feature maps and $W_t$ respectively. We ignore the subscription $t$ for simplicity. 
Here ${\mathbf W}$ is in fact a channel-wise projection matrix.

We can further reduce the computation by simplifying the ${\mathbf W}_t$ to be a channel-wise scaling vector ${\mathbf W}_t$ with size $C$ as illustrated in Figure~\ref{fig.network1}. 

Formally, ${\mathbf W} = \{w_c\}, c \in \{1,\ldots,C\}$.
\begin{equation}
x^\prime_{mnc} = x_{mnc}*w_{c},
\end{equation} where $x^\prime_{mnc}$ and $x_{mnc}$ denotes elements from input and output feature maps respectively. 

\begin{figure*}
\centering
\includegraphics[width = 0.56\linewidth]{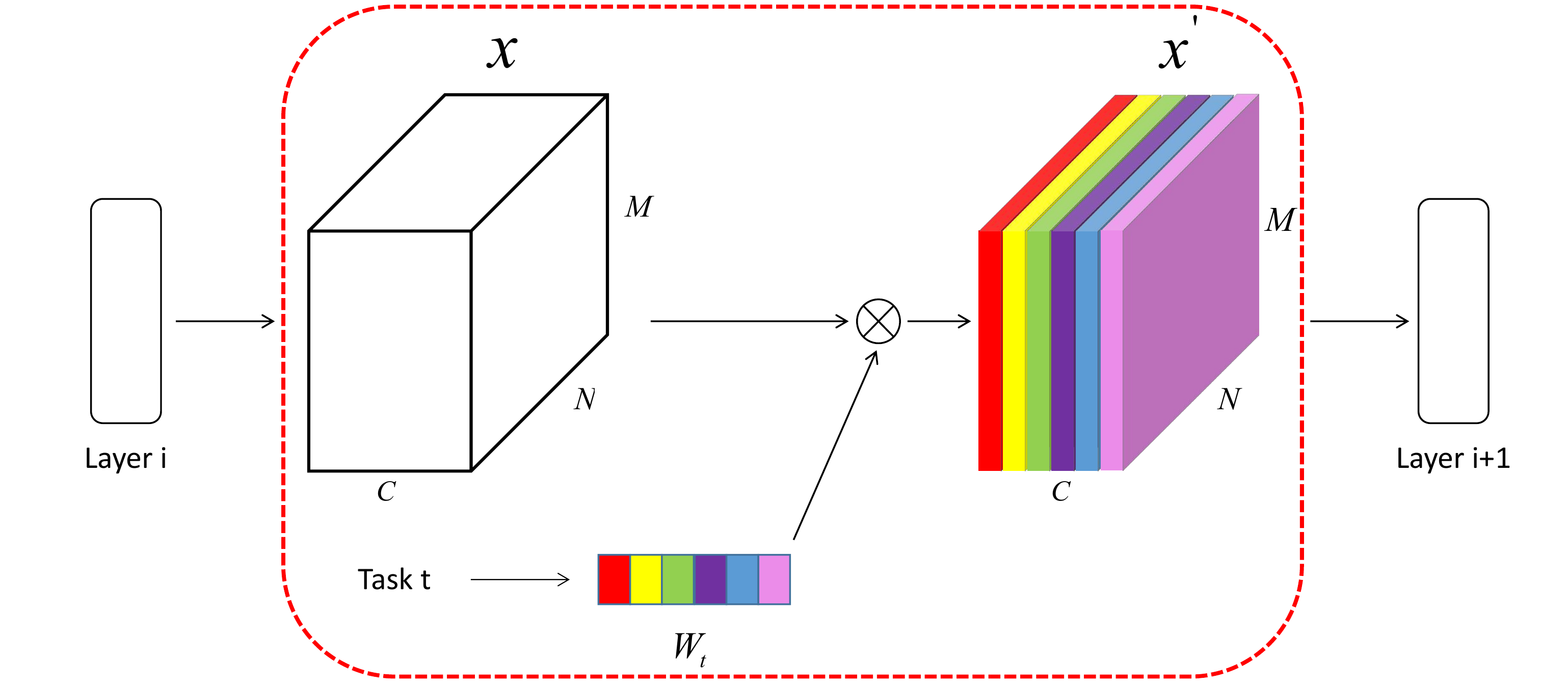}
\caption{Structure of the proposed Modulation Module which adapts features via learned weights with respect to each task. This module can be inserted between any layers and maintain the network structure.}
\label{fig.module}
\end{figure*}

Compared with the channel-wise scaling vector design, we observe empirically the overall improvement from the channel-wise projection matrix design is marginal, hence we will mainly discuss and evaluate the simpler channel-wise scaling vector option. This module can be easily implemented by adding task specific linear transformations as shown in Figure~\ref{fig.module}.




\subsection{Training}

The modulation parameters ${\mathbf W}_t$ are learned together with the neural network parameters through back-propagation. In this paper, we use triplet loss~\cite{schroff2015facenet} as the objective for optimization. More specifically, given a set of triplets from different tasks $(I_a,I_p,I_n,t) \in {\mathbf T}$, 
\begin{eqnarray}
L &=& \sum_{{\mathbf T}} [\|f_a - f_p\|^2 + \alpha - \|f_a - f_n\|^2)]_{+} \\
f_{a,p,n} &=& F(I_{a,p,n} | \theta, {\mathbf W}_t))
\end{eqnarray}, where $\alpha$ is the expected distance margin between positive pair and negative pair, $I_a$ is the anchor sample, $I_p$ is the positive sample, $I_n$ is the negative sample and $t$ is the task.

When training the Neural Network with a discriminative loss, we argue that by introducing the Modulation module into the neural network, it will learn to leverage the additional knobs to decouple unrelated tasks and couple related ones to minimize the training loss. In the toy experiment shown in Table~\ref{tab:three}, we primarily show that our method can surpass fully independent learning. The reduced ratios of conflicting mini-batches in training as shown in Table~\ref{tab:three} also validate our design.

The learned ${\mathbf W}_*$ capture the relationship of tasks implicitly. We obtained ${\mathbf W}_s$, ${\mathbf W}_y$ and ${\mathbf W}_o$ for {\em smile}, {\em young}, {\em open-mouth} respectively. Then the element-wise difference between ${\mathbf W}_s$ and ${\mathbf W}_o$,$\nabla {\mathbf W}_{s,o}$, and the difference between ${\mathbf W}_s$ and ${\mathbf W}_y$, $\nabla {\mathbf W}_{s,y}$, are obtained to measure their relevancy. The mean and variance of $\nabla {\mathbf W}_{s,o}$ is 0.18 and 0.03 while the mean and variance of  $\nabla {\mathbf W}_{s,y}$ is 0.24 and 0.047. 

We further empirically validate this assumption by introducing an additional regularization loss to encode human prior knowledge on the tasks' relevancy. We assume the learned ${\mathbf W}$ for {\em smile} would be more similar to the one for {\em open mouth} compared with the one for {\em young}. We regularize the pairs of relevant tasks to have similar task-specific ${\mathbf W}$s with 
\begin{equation}
L_a = max(0, \|{\mathbf W}_i - {\mathbf W}_j\|^2 + \beta - \|{\mathbf W}_i - {\mathbf W}_k\|^2)
\end{equation}, where $\beta$ is the expected margin, $i,j,k$ denotes three tasks, and task pair $(i,j)$ is considered more relevant compared to task pair $(i,k)$. $L_a$ is weighted by a hyper-parameter $\lambda$ and combined with the above triplet loss over samples in training.

As shown in Table~\ref{tab:three}, the accuracy of our method augmented with this regularization loss is better but the gap is only marginal. This suggests that without encoding prior knowledge through the loss, the learned ${\mathbf W}$s may implicitly capture task relationships in a similar way. On the other hand, it is impractical to manually define all pairwise relationships when the number of tasks scales up, hence we ignore this regularization loss in our large-scale experiments.

\section{Experiments}
\begin{table}
	\begin{center}
        \begin{tabular}{|c|c|c|}
        \hline
        Name & Operation  & Output Size \\
        \hline
        \hline
        conv1 & $3 \times 3$ convolution &$148 \times 148 \times 32$\\
        block2  & Conv-Pool-ResnetBlock &$73 \times 73 \times 64$ \\
        block3  & Conv-Pool-ResnetBlock &$35 \times 35 \times 128$ \\
        block4  & Conv-Pool-ResnetBlock &$16 \times 16 \times 128$ \\
        block5  & Conv-Pool-ResnetBlock &$7 \times 7 \times 128$ \\
         fc     & Fully-Connected &$ 256$\\
          \hline
        \end{tabular}
	\end{center}
    \caption{Our Basic Neural Network Architecture: Conv-Pool-ResnetBlock stands for a $3\times 3$ conv-layer followed by a stride 2 pooling layer and a standard residual block consist of 2 $3\times 3$ conv-layers.}
    	\label{tab:basenet}
\end{table}

In the experiments, we evaluate the performance of our approach on the face retrieval and product retrieval tasks.

\subsection{Setup~\label{sec:expsetup}}

In both retrieval settings, we define a task as retrieval based on a certain attribute of either face or product. Both datasets have the per-image annotation for each attribute. To quantitatively evaluate the methods under the retrieval setting, we randomly sample image triplets from their testing sets as our benchmarks. Each triplet consists of an anchor sample $I_a$, a positive sample $I_p$, and a negative sample $I_n$. Given a triplet, we retrieve one sample from $I_p$ and $I_n$ with $I_a$ and consider it a success if $I_p$ is preferred. In our method, we extract discriminative features with the proposed network and measure image pair distance by their euclidean distance of features. The accuracy metric is the ratio of successfully retrieved triplets.

Unless stated otherwise, we use the neural network architecture in Table~\ref{tab:basenet} for our method, our re-implementation of other state-of-the-art methods, and our baseline methods.

We add the proposed Modulation modules to all layers from block4 to the final layer and use ADAGRAD~\cite{duchi2011adaptive} for optimization in training with learning rate $0.01$. We uniformly initialize the parameters in all added modules to be 1. We use the batch size of $180$ for 20 tasks and $168$ for 7 tasks joint training. 
In each mini-batch, we evenly sample triplets for all tasks. Our method generally converges after 40 epochs.
\subsection{Face Retrieval}
\subsubsection{Dataset}
\begin{table}
	\begin{center}
   \resizebox{12cm}{6cm}{\begin{tabular}{|c|c|c|c|c|c|c|c|}
            \hline
            Methods: & Ours & CSN & \makecell{ITN} & \makecell{FSN}  & \makecell{IB-256} & \makecell{IB-25} & \makecell{Only mask} \\
            \hline
            \hline
            \makecell{Average \\ Accuracy} & \textbf{84.86\%} & 72.81\% & 84.61\% & 69.4\% & 83.69\% & 75.47\% & 76.32\% \\
            \hline
            \makecell{Number of \\ Baseline Parameters} & $3M $ & $3M $ &$3M $ & $3M$ & $3M$ & $3M$ & $3M$\\
            \hline
             \makecell{Number of \\ additional Parameters} &  $10k$ & $3k$ &$ 51M $ & $0$ & $1.3M$ & $128k$ &$10k$\\
            \hline
            \hline
            smile & \textbf{93.77\%} & 75.59\% & 93.32\% & 78.83\% & 92.76\% & 82.91\% & 87.64\%\\
            \hline
            shadow & \textbf{94.67\%} & 92.83\% & 92.25\% & 85.39\% & 92.83\% & 88.02\% & 86.41\% \\
            \hline
            bald & \textbf{91.83\%} & 87.80\% & 90.70\% & 81.79\% & 89.47\% & 78.11\% & 88.42\% \\
            \hline
            are-eyebrows & 78.36\% & 63.94\% & \textbf{79.60\%} & 66.19\% & 76.84\% & 66.00\% &72.10\%\\
            \hline
            chubby & \textbf{90.2\%} & 85.32\% & 87.29\% & 79.06\% & 88.66\% &82.79\% & 85.39\%\\
            \hline
            double-chin & \textbf{91.45\%} & 85.61\% & 89.57\% & 81.15\% & 89.92\% & 83.08\% & 87.19\%\\
            \hline
            high-cheekbone & 88.53\% & 71.25\% & \textbf{88.93\%} & 74.57\% & 87.25\% & 76.53\% & 82.80\%\\
            \hline
            goatee & \textbf{94.47\%} & 90.66\% & 94.06\% & 83.48\% & 94.17\% & 84.68\% & 91.52\%\\
            \hline
            mustache & \textbf{93.41\%} & 89.21\% & 93.23\% & 82.40\% & 93.21\% & 87.52\% & 89.89\%\\
            \hline
            no-beard & \textbf{93.84\%} & 82.35\% & 93.69\% & 80.52\% &93.98\% & 86.51\% & 85.69\%\\
            \hline
            sideburns & 95.27\% & 90.95\% & 94.88\% & 86.20\% & 95.04\% &88.81\% &91.85\%\\
            \hline
            bangs & \textbf{90.22\%} & 71.91\% & 89.96\% & 69.96\% & 89.13\% & 78.75\% &80.34\%\\
            \hline
            straight-hair & 72.98\% & 63.31\% & \textbf{73.24\% }& 61.70\%& 71.98\% & 62.33\% & 65.47\% \\
            \hline
           wavy-hair& \textbf{76.59\% }& 59.34\% & 76.10\% & 59.49\% &75.62\% & 64.04\%  &65.11\%\\
            \hline
            receding-hairline & \textbf{87.33\%} & 75.63\% & 86.93\% & 72.02\% & 86.24\% & 80.17\% & 79.94\%\\
            \hline
            bags-eyes & 85.90\% & 76.39\% & \textbf{85.93\%} & 72.39\% & 84.64\% & 76.01\%& 82.05\%\\
            \hline
            bushy-eyebrows & \textbf{88.73\%} & 79.22\% & 88.32\% & 74.52\% & 88.44\% & 80.50\% & 80.50\% \\
            \hline
            young & \textbf{84.87\%} & 60.61\% & \textbf{84.90\%} & 61.55\% & 83.48\% &  73.05\%& 66.23\%\\
            \hline
            oval-face & \textbf{72.21\%} & 64.33\% & 71.52\% & 63.54\% &70.16\% & 62.10\%  &65.10\%\\
            \hline
            mouth-open & \textbf{94.59\%} & 87.32\% & 94.40\% & 72.71\% & 92.22\% & 89.03\% & 86.59\%\\
             \hline
        \end{tabular}}
	\end{center}
    \caption{Accuracy comparison on the joint training of 20 face attributes: with far fewer parameters, our method achieves best mean accuracy over the 20 tasks compared with the competing methods.}
	\label{tab:twenty}
\end{table}
We use Celeb-A dataset~\cite{liu2015faceattributes} for the face retrieval experiment. Celeb-A consists of more than 200,000 face images with binary annotations on 40 face attributes related to age, expression, decoration, etc. We select 20 attributes more related to face appearance and ignore attributes around decoration such as eyeglasses and hat for our experiments. We also report the results on 40 attributes to verify the effectiveness on 40 attributes. 
\renewcommand{\arraystretch}{0.95}

We randomly sampled 30000 triplets for training and 10000 triplets for testing for each task. Our basic network architecture is shown in Table~\ref{tab:basenet}. We augment it by inserting our gradient modulation modules and train from scratch.
\begin{table}
	\begin{center}
 \scalebox{0.95}{\begin{tabular}{|c|c|c|c|c|c|c|c|c|}
			\hline
            & smile & ovalface & shadow & bald & arc-eyebrows & big-lips & big-nose \\
			\hline
			smile & -  & {\color{red}51.56}/48.47 & {\color{red}67.70}/26.33 & {\color{red}67.82}/32.30 & {\color{red}52.32}/45.40 & {\color{red}54.83}/49.49 & {\color{red}58.72}/45.25\\
			\hline
            ovalface & {\color{red}51.56}/48.47& - & {\color{red}67.36}/26.94 & {\color{red}64.99}/35.29 &{\color{red}57.86}/50.13 & {\color{red}57.74}/49.32& {\color{red}54.98}/46.64 \\
			\hline
            shadow & {\color{red}67.70}/26.33 & {\color{red}67.36}/26.94 & -  & {\color{red}91.67}/30.54 & {\color{red}66.87}/26.48 & {\color{red}72.51}/28.25 &{\color{red}69.90}/29.99  \\
			\hline
            bald & {\color{red}67.82}/32.30& {\color{red}64.99}/35.29 & {\color{red}91.67}/30.54 & - & {\color{red}61.74}/31.67 & {\color{red}67.60}/36.22 & {\color{red}72.66}/41.04 \\
			\hline
            arc-eyebrows & {\color{red}52.32}/45.40  & {\color{red}57.86}/50.13 & {\color{red}66.87}/26.48 & {\color{red}61.74}/31.67 & -& {\color{red}58.86}/51.13 & {\color{red}50.34}/41.43 \\
            \hline
            big-lips & {\color{red}54.83}/46.49  & {\color{red}57.74}/49.32& {\color{red}72.51}/28.25 & {\color{red}67.70}/36.22 & {\color{red}58.86}/51.13 & - & {\color{red}55.20}/46.84 \\
                        			\hline
            big-nose & {\color{red}58.72}/45.25  & {\color{red}54.98}/46.64 & {\color{red}69.90}/29.99 & {\color{red}72.66}/41.04 & {\color{red}50.34}/41.43&  {\color{red}55.20}/46.84& - \\            
			\hline
		\end{tabular}}
	\end{center}
    \caption{Comparison of UCR between different tasks on joint training of seven face attributes with our method (red) and the fully shared network baseline (black): we quantitatively demonstrate the mitigation of {\em destructive interference} with our method.}
	\label{tab:seven-gradient}
\end{table}

\subsubsection{Results~\label{sec:faceresults}}



We report our evaluation of the following methods in Table~\ref{tab:twenty}:
\begin{itemize}
    \item Ours: we insert the proposed Modulation modules to the block4, block5, and fc layers to the network in Table~\ref{tab:basenet} and jointly train it with all training triplets from 20 tasks;
    \item Conditional Similarity Network (CSN) from Veit et al.~\cite{Veit2017}: we follow the open-sourced implementation from the authors to replace the network architecture with ours and jointly train it with all training triplets from 20 tasks;
    \item Independent Task-specific Network(ITN): in this strong baseline we train 20 task-specific neural networks with training triplets from each task independently;
    \item Single Fully-shared Network(FSN): we train one network with all training triplets.
    \item Independent Branch 256(IB-256): based on shared parameters, we add task-specific branch with feature size 256.
    \item Independent Branch 25(IB-25): based on shared parameters, we add task-specific branch with feature size 25.
    \item Only-mask: our network is pretrained from the independent branch model, the shared parameters are fixed and only the module parameters are learned. 
\end{itemize}
\renewcommand{\arraystretch}{1.1}
\begin{table}
	\begin{center}
	 \scalebox{0.95}{	\begin{tabular}{|c|c|c|c|c|c|c|c|c|}
			\hline
            Face Attributes: & smile & ovalface & shadow & bald & arc-eyebrows & big-lips & big-nose& \makecell{Average\\ Accuracy} \\
			\hline\hline
            \makecell{Single \\Fully-shared Network} & 78.39\%  & 64.39\% & 79.55\% &77.62\% & 69.17\%& 61.71\% & 68.88\% & 71.38\% \\
			\hline
            \makecell{Independent \\Task-specific Networks} &93.32\% & 71.52\%  & 92.25\% & 90.70\%&79.60\% & \textbf{67.35\%}& 84.35\% & 82.72\% \\
			\hline
			CSN & 91.39\%  & 68.41\% &92.51\% &\textbf{90.79\%} &77.53\% &65.79\% & 82.03\% & 81.20\%\\
			\hline
            Ours (from block5) &93.35\%& 70.47\% & 90.44\% &88.79\% &77.12\% & 66.36\%& 83.84\% & 81.48\%\\
			\hline
            Ours (from block4)&93.69\% & 71.44\% & 92.06\%  & 90.66\% &\textbf{80.00\%} &67.15\% &84.26\%  &82.75\% \\
			\hline
            Ours (from block3) &93.83\% & 71.04\% & \textbf{93.28\%} & 90.66\% &79.76\% & 67.53\% & \textbf{84.76\%} &\textbf{82.98\%}\\
			\hline
            Ours (from block2)& \textbf{94.11\%}  & 71.94\% &92.5\% &90.70\% &78.66\% &66.36\% & 84.10\% &82.62\% \\
			\hline
			\hline
            \makecell{channel-wise projection\\ (from block4)} & \textbf{94.10\%}  & \textbf{71.98\%} &92.69\% &90.58\% &78.95\% &66.78\% & 84.48\% & 82.79\%\\
			\hline
		\end{tabular}}
	\end{center}
    \caption{Ablation Study of our method: with more layers modulated by the proposed method, performance generally improves; channel-wise projection module is marginally better than the default channel-wise scaling vector design.}
	\label{tab:seven}
\end{table}
Single Fully-shared network and CSN severely suffer from the {\em destructive interference} as shown in Table~\ref{tab:twenty}. Note when jointly training only 7 tasks, CSN performs much better than the fully-shared network and similarly to fully shared network with additional parameters as shown in Table~\ref{tab:seven}. However, it does not scale up to handle as many as 20 tasks. Since the majority of the parameters are naively shared across tasks until the last layer, CSN still suffers from {\em destructive interference}. 


We then compare our methods with Independent Branch methods. Independent Branch methods naively add task specific branches above the shared parameters.  The branching for IB-25 and IB256
begins at the end of the baseline model in Table 2, i.e., different attributes
have different branches after the FC layer. As illustrated in Table~\ref{tab:twenty}, our method clearly outperforms them with much fewer task-specific parameters. Regarding the number of additional parameters, we observe that to approximate accuracy of our method, this baseline needs about $1.3M$ task-specific parameters, which is $100$ times of ours. The comparison indicates that our module is more efficient in leveraging additional parameters budget.

\begin{figure*}[h]
\centering
\subfigure[Smile]{
\includegraphics[width=0.9\linewidth]{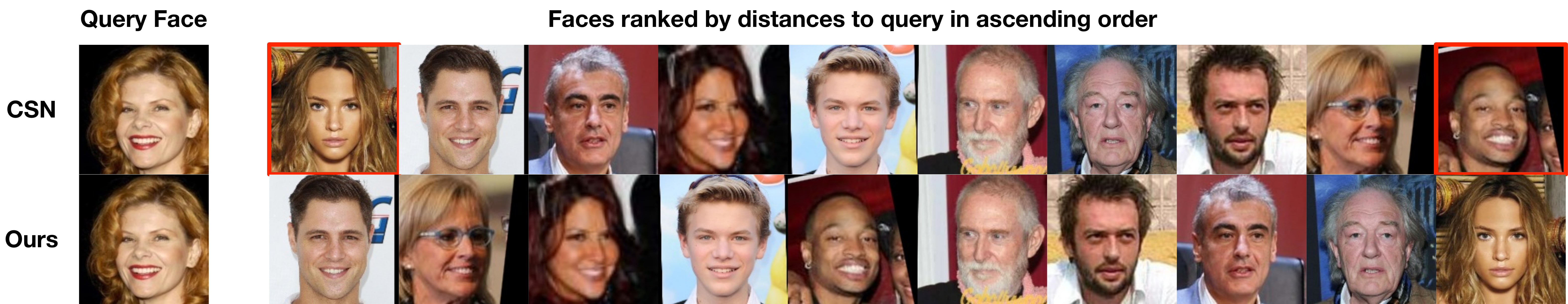}
}
\subfigure[Sideburns]{
\includegraphics[width=0.9\linewidth]{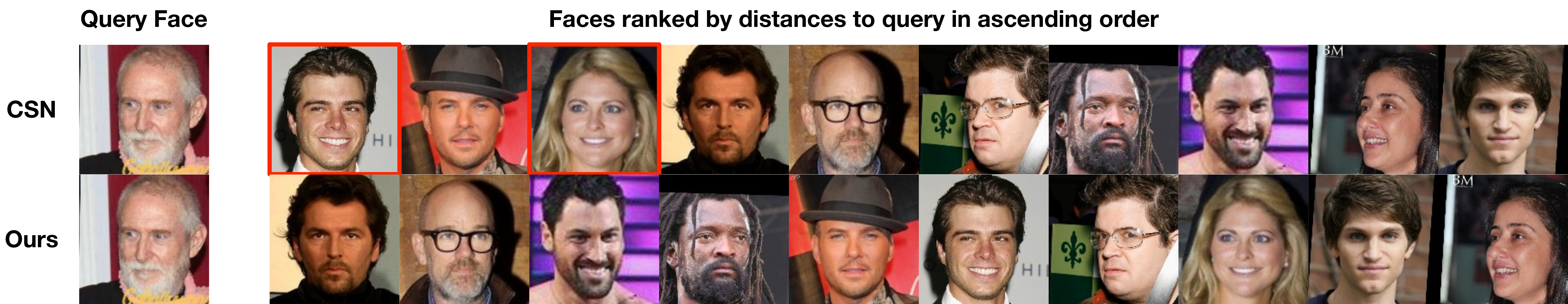}
}
\caption{Example face retrieval results in two tasks: using models jointly trained for 20 face attributes with CSN and our method respectively. Some incorrectly ranked faces are highlighted in red.}
\label{fig.qual}
\end{figure*}

\renewcommand{\arraystretch}{1.1}
\begin{table}
	\begin{center}
		\begin{tabular}{|c|c|c|c|c|c|}
			\hline
			Tasks: & class & closure &gender & heel & \makecell{Average Accuracy} \\
			\hline\hline
            \makecell{Single Fully-shared Network} & 78.95\%  &80.33\%  & 69.22\% & 73.35\% & 75.46\% \\
			\hline
            \makecell{Independent Task-specific  Networks}  &92.01\%  &89.12\%  &79.10\% &85.97\%  & 86.61\% \\
			\hline
            CSN~\cite{Veit2017} &93.06\%  & 89.37\% & 78.09 & 86.42\% & 86.73\% \\ 
			\hline
			Ours  & \textbf{93.34\%}  & \textbf{90.57\%}  &\textbf{79.50\%}  & \textbf{89.27\%} & \textbf{88.17\%}\\
			\hline
		\end{tabular}
	\end{center}
    \caption{Accuracy Comparison on joint training of 4 product retrieval tasks on UT-Zappos50k: our method significantly outperforms others.}
	\label{tab:shoes_table}
\end{table}
Compared with the independently trained task-specific networks, 
our method achieves slightly better average accuracy with almost 20 times fewer parameters. Notably, our method achieves obvious improvement for both face shape related attributes ({\em chubby}, {\em double chin}) and all three beard related attributes ({\em goatee}, {\em mustache}, {\em sideburns}), which demonstrates that the proposed method does not only decouple unrelated tasks but also adaptively couples related tasks to improve their learning. We show some example retrieval results in Figure~\ref{fig.qual}. 

We reported the Update Compliance Ratio(UCR) comparison in Table~\ref{tab:seven-gradient}. Our method significantly improves the UCR in the joint training for all task pairs. This indicates that the proposed module is effective in alleviating the {\em destructive interference} by leading the gradients over shared parameters from different tasks to be more consistent. 


To further validate that the source of improvement is from better shared parameters instead of simply additional task specific parameters. We keep our shared parameters fixed as the ones trained with the strong baseline IB-256 and only make the modulation modules trainable.
As reported in the last column in Table~\ref{tab:twenty}, the results are not as good as our full pipeline, which suggests that the proposed modules improved the learning of shared parameters. To validate the effectiveness of our method on 40 attributes, we evaluate our method on 40 attributes and obtain average 85.75\% which is significant better than 78.22\% of our baseline IB-25 which has same network complexity but with independent branches.


\subsubsection{Ablation Study~\label{exp:ablationstudy}}
In Table~\ref{tab:seven}, we evaluate how the performance evolves when we insert more  Modulation modules into the network. By adding proposed modules to all layers after block$N$, $N = {5,4,3,2}$, we observe that the performance generally increases with more layers modulated. This is well-aligned with our intuition that with gradients modulated in more layers, the destructive inference problem gets solved better. Because early layers in the neural networks generally learn primitive filters~\cite{zeiler2014visualizing} shared across a broad spectrum of tasks, shared parameters may not suffer from conflicting updates. Hence the performance improvement saturates eventually.

We also experiment with channel-wise projection matrix instead of channel-wise scaling vector in the proposed modules as introduced in Section~\ref{sec:GM}. 
We observe marginal improvement with the more complicated module, as shown in the last row of Table~\ref{tab:seven}. This suggests that potentially with more parameters being modulated, the overall performance improves at the cost of additional task-specific parameters.
It also shows that the proposed channel-wise scaling vector design is a cost-effective choice.


\subsection{Product Retrieval}

\subsubsection{Dataset}

We use UT-Zappos50K dataset~\cite{finegrained,semjitter} for the product retrieval experiment. UT-Zappos50K is a large shoe dataset consisting of more than 50,000 catalog images collected from the web. 
The datasets are richly annotated and we can retrieve shoes based on their type, suggested gender, height of their heel, and the closing mechanism. We jointly learn these 4 tasks in our experiment.
We follow the same training, validation, and testing set splits as Veit et al.~\cite{Veit2017} to sample triplets. 
\subsubsection{Results}

As shown in Table~\ref{tab:shoes_table}, our method is significantly better than all other competing methods. 
Because CSN manually initializes the 1-dimensional mask for each attribute to be non-overlapping, their method does not exploit their correlation well when two tasks are correlated. We argue that naively sharing features for all tasks may hinder the further improvement of CSN due to gradient discrepancy among different tasks. In our method, proposed modules are inserted in the network and the correlation of different tasks are effectively exploited. Especially for {\em heel} task, our method obtains a nearly 3 point gain over CSN.  
Note that because our network architecture is much simpler than the one used by Veit et al.~\cite{Veit2017} and does not pre-train on ImageNet. The numbers are generally not compatible to those reported in their paper.

\section{Discussion}

\subsection{General applicability}

In this paper, we mainly discuss multi-task learning with application in image retrieval in which each task has similar network structure and loss functions. 
By design the proposed module is not limited to a specific loss and should be applicable to handle different tasks and different loss functions.

In general multi-task learning, each task may have its specifically designed network architecture and own loss, such as face detection and face alignment~\cite{ranjan2017all,zhang2016joint}, learning object detection and segmentation~\cite{He_2017_ICCV,girshick14CVPR}, learning semantic segmentation and depth estimation~\cite{mousavian2016joint,wang2015towards}. 
The signals from different tasks could be explicitly conflicting as well and lead to severe \textit{destructive interference} especially when the number of jointly learned tasks scale up. When such severe \textit{destructive interference} happens, the proposed module could be added to modulate the update directions as well as task-specific features. We leave it as our future work to validate this assumption through experiments.

\subsection{Speed and Memory size Trade-off}

Similar to a multi-branch architecture and arguably most multi-task learning frameworks, our method shares the problem of runtime speed and memory size trade-off in inference. One can either choose to keep all task-specific feature maps in memory to finish all the predictions in a single pass or iteratively feed-forward through the network from the shared feature maps to keep a tight memory foot-print.
However, we should highlight that our method can achieve better accuracy with a more compact model in storage. Either a single pass inference or iterative inference could be feasible with our method. Since most computations happen in the early stage in inference, with the proposed modules, our method only added 15\% overhead in feed-forward time. The feature maps after block4 are much smaller than the ones in the early stages, so the increased memory footprint would be sustainable for 20 tasks too.


\section{Conclusion}

In this paper, we propose a Modulation module for multi-task learning. We identify the {\em destructive interference} problem in joint learning of unrelated tasks and propose to quantify it with Update Compliance Ratio. The proposed modules alleviate this problem by modulating the directions of gradients in back-propagation and help extract better task-specific features by exploiting related tasks. 
Extensive experiments on CelebA dataset and UT-Zappos50K dataset verify the effectiveness and advantage of our approach over other multi-task learning methods.

\subsubsection*{Acknowledgements}
This work was supported in part by National Science Foundation grant IIS-1217302, IIS-1619078, the Army Research Office ARO W911NF-16-1-0138, and Adobe Collaboration Funding. 

\pagebreak


\bibliographystyle{splncs04}
\bibliography{egbib.bib}

\end{document}